\documentclass{article}

\usepackage[preprint, nonatbib]{neurips_2020}
\usepackage[square, numbers]{natbib}

\usepackage[table]{xcolor}
\definecolor{greenbars}{HTML}{33CC66}
\definecolor{yellowbars}{HTML}{FFD700}
\definecolor{redbars}{HTML}{FF8080}

\usepackage[utf8]{inputenc} %
\usepackage[T1]{fontenc}    %
\usepackage[hyphens,spaces,obeyspaces]{url}           
\usepackage[colorlinks=true, citecolor=black, linkcolor=black, urlcolor=black, breaklinks=true]{hyperref}       %
\usepackage{booktabs}
\usepackage{todonotes}%
\usepackage{amsfonts}       %
\usepackage{nicefrac}       %
\usepackage{microtype}      %
\usepackage{newfloat}
\usepackage{multirow}
\usepackage{subcaption}
\usepackage{listings}
\usepackage{amsmath, amssymb}

\usepackage{siunitx}

\definecolor{jccolor}{rgb}{0.1,0.7,0.8}
\definecolor{vlcolor}{rgb}{0.9,0.1,0.1}
\definecolor{cpcolor}{rgb}{0.3,0.3,0.7}
\definecolor{ascolor}{RGB}{163,96,50}
\definecolor{ahcolor}{rgb}{0.36, 0.54, 0.66}

\title{Tabular Representation, Noisy Operators, and Impacts on Table Structure Understanding Tasks in LLMs}

\author{
  Ananya Singha \\
  Microsoft, India \\
  \texttt{t-asingha@microsoft.com} \\
  \And
  José Cambronero \\
  Microsoft, USA \\
  \texttt{jcambronero@microsoft.com} \\
  \And
  Sumit Gulwani \\
  Microsoft, USA \\
  \texttt{sumitg@microsoft.com} \\
  \And
  Vu Le \\
  Microsoft, USA \\
  \texttt{levu@microsoft.com} \\
  \And
  Chris Parnin \\
  Microsoft,USA \\
  \texttt{chrisparnin@microsoft.com} \\
}

\begin{document}

\maketitle

\begin{abstract}
  Large language models (LLMs) are increasingly applied
  for tabular tasks using in-context learning.
  The prompt representation for a table may
  play a role in the LLMs ability to process the table.
  Inspired by prior work, we generate a
  collection of
  self-supervised table structure understanding tasks (e.g. navigate to
  a cell and row; transpose the table) and evaluate
  the performance differences when using eight formats.
  In contrast to past work, we introduce eight noise operations
  inspired by real-world messy data and
  adversarial inputs, and show that these
  can impact LLM performance across formats for
  different structural understanding tasks.

\end{abstract}

\section{Introduction}

Recent progress in large language models (LLMs) has 
enabled substantial gains when performing data-related tasks, such as table question answering~\cite{chen2023large}, semantic type annotation~\cite{Suhara:2022}, and
data wrangling~\cite{semantictypes:2023}---often with just
in-context learning.
However, data is also messy. Tabular data often arrives in a semi-structured format, with inconsistent shapes, missing entries, and unnormalized or inconsistently formatted values. This makes the task of processing and understanding tabular data particularly challenging for any system, including LLMs. For example, based on product telemetry, 21\% of Excel files imported using LLM-based Dataverse Copilot\footnote{\url{https://powerapps.microsoft.com/en-us/blog/introducing-an-easier-than-ever-experience-to-import-data-from-excel/}} were missing headers. Furthermore, in industrial settings, challenges such as privacy, compliance, and even potentially
adversarial table inputs constrain how data can be handled and processed~\cite{cahoon2022need}.

In this work we systematically explore the impact that the tabular representation format and real-world-inspired noise have on LLMs' ability to perform 
basic structural table understanding tasks~\cite{sui2023evaluating} through in-context learning. 
Like prior work, we generate
self-supervised structural table tasks to assess
structural understanding. In contrast to prior
investigations, we incorporate eight noise-inducing
operations---such as renaming columns or transposing the table---that manipulate the table's structure
in ways that emulate messy data~\cite{microsoft-data-cleaning} 
or even adversarial inputs.

We evaluate both fact-finding and transformation tasks
over seven public datasets,
eight table
representations commonly used in data-science and eight noise-inducing
operations. In contrast to prior work, 
we find that HTML does not 
seem to provide the best performance at fact-finding
or transformation tasks. We find that a dataframe-based
format (DFLoader) obtains the highest overall
pass@1 (79.79\%) in fact-finding tasks and the highest
overall F1 score (98.55\%) for transformation tasks.
We find that applying noise operations to tables can affect
performance
in fact-finding and transformation
tasks. For example, introducing semi-structured
content impacts data type detection (e.g. 
JSON format's pass@1 drops by 12.43\%) and introducing
sequential column naming can degrade performance for
a column reordering task (e.g. comma-separated-value
format's F1 score degrades by 67.33\%).

We believe 
future work can build on our findings by 
exploring the extent to which
these structural table understanding tasks
relate to downstream task performance. Furthermore,
such work should include (and extend) our noise
operations to evaluate the impact of structural changes.

In summary, our key contributions are:
\begin{enumerate}
    \item Extending self-supervised table structure understanding tasks by incorporating noise operations inspired by real-world noise
    \item An extensive evaluation over eight table formats and eight noise-inducing operations
    \item Our data and code to facilitate
    future work on
    structural table understanding
\end{enumerate}

\section{Related Work}

Transformer architectures have 
led to state-of-the-art performance
in NLP and other areas of
machine learning.
This has motivated a line of research
developing transformer models
designed for tabular tasks, 
such as table question answering.
These models (e.g. TUTA~\cite{wang2021tuta}, TAPAS~\cite{herzig2020tapas}) are predominantly developed by training and fine-tuning on large corpus of data scraped  from Wikipedia and introducing different
attention mechanisms.
Prior work~\cite{koleva2022analysis}
has carried out a detailed analysis of
how these mechanisms work and
how they affect table understanding tasks.
In contrast,
we focus on a general LLM (GPT3) rather than
ones designed for table tasks and carry out
our experiments using in-context learning.
Importantly, we scope our experiments to
understand the impact of table
representation format (subject to noise operations) on self-supervised
table structure tasks.

Prior work has used in-context learning to carry out tasks
on tabular data. For example,
TableLLM showed that LLMs can perform classification tasks over tabular datasets. Techniques like
chain-of-thought~\cite{wei2022chain} have
been further refined in the context of tabular data~\cite{zha2023tablegpt, chen2022large}. In contrast, 
we focus on self-supervised
table structure tasks and consider the impact of table formats and the robustness
to noise inspired by
real-world data issues
and adversarial behaviors.

The closest related work is~\cite{sui2023evaluating}, which examines
LLM performance on structural understanding tasks as a function
of different tabular formats. 
Our work extends this line of research with other
formats, new fact-finding and transformation tasks, and noise
operations inspired
by messy data.

\section{Methodology}
To evaluate the extent to which different table representation formats and noise operations affect an LLM's ability to correctly
answer structural table understanding tasks, we 
generate a collection of
self-supervised tasks (i.e. where we can derive the task and answer from the table without the need for annotation). We now describe this approach in detail.

Let $T$ be a flat table
with a header, $F$
be a set of table representation formats (e.g. JSON),
each of which
transforms $T$ into
a corresponding string
representation for the prompt. Let
$N$ be the set of noise operations (e.g. shuffle header names), each of which
transforms $T$ into $T'$.
Let $Q$ be the set of
self-supervised tasks (e.g. lookup the value at row X and column Y),
each of which given
a table generates
a collection $\{ ( t, a ) \}$ of self-supervised task question $t$
and answer $a$ pairs. 
We create an evaluation benchmark for a given table
of the form $\{ (f(n(T)), t, a) |\ q \in Q, f \in F, n \in N, (t, a) \in q(f(n(T))) \}$. For each 
$(t, a)$, we then compare $a$ to the LLM's
answer given $t$ and $f(n(T))$.

\subsection{Table and Formats}

\noindent{\textbf{Tables}}: We scope our experiments
to flat tables with
a header row. Furthermore,
each column must contain
a single datatype (e.g. string, numeric, date).

\begin{figure*}[hbt]
    \centering
    \includegraphics[width=\textwidth]{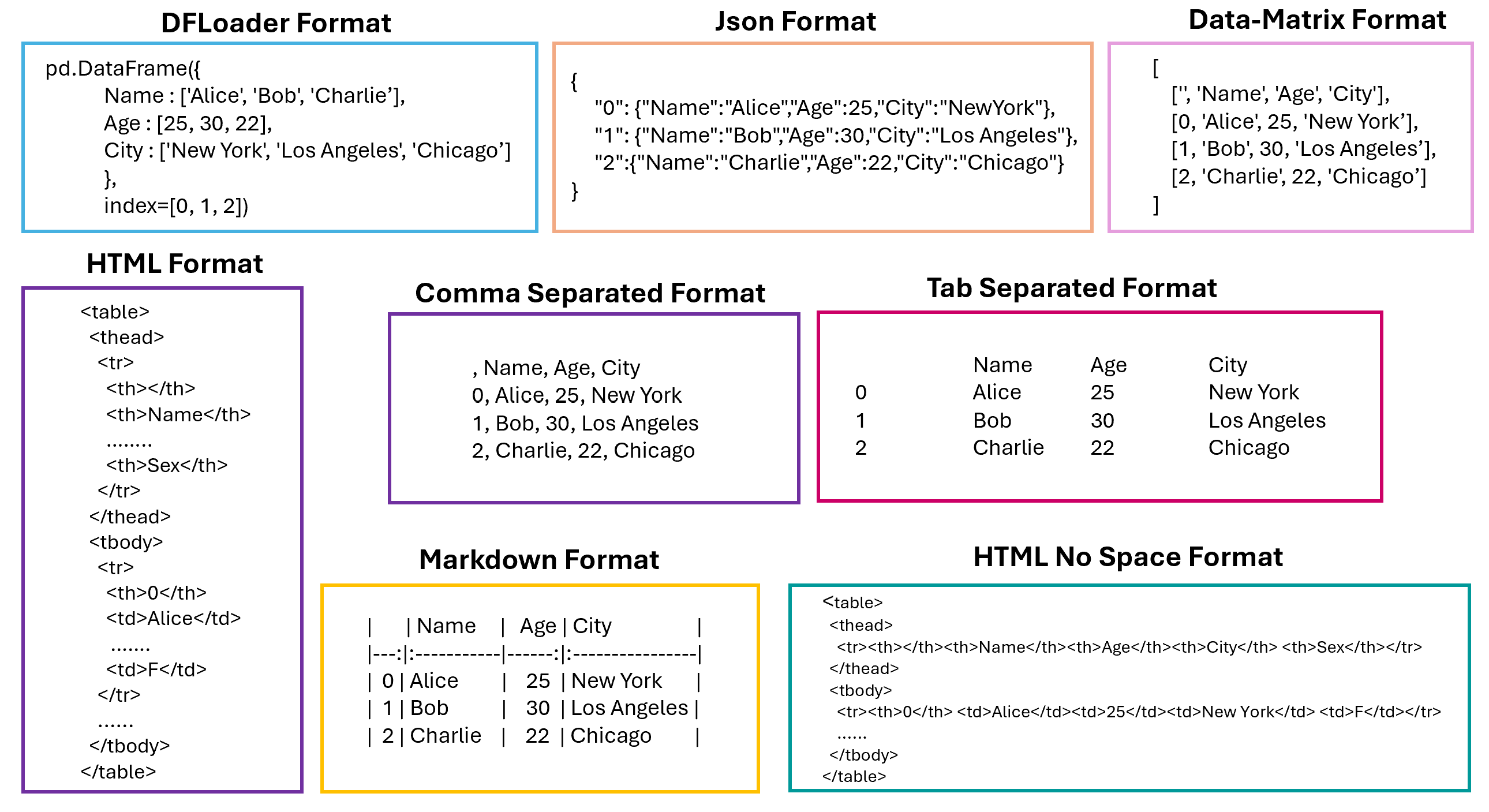}
    \caption{
    Our evaluation considers 8
    different table representation formats that are popular in the
    data science domain.
    }
    \label{fig:tabformats}
\end{figure*}
\noindent{\textbf{Formats}}: We represent these
tables with the following
8 popular formats, summarized in
Figure~\ref{fig:tabformats}:
DFLoader, JSON, Data-Matrix,
Markdown, Comma-Separated-Values,
and Tab-Separated-Values format.
DFLoader corresponds to the associated
Python code snippet to define the
table using the Pandas
DataFrame API. Data-Matrix
format represents each row as
a mixed-type list of values. HTML format represent the table using nested tags. HTML No Space inlines HTML by removing whitespaces.
Note that our tables include both
headers and row indices.

\subsection{Noise Operations}
We
explore the extent
to which noise operations
can impact the LLM's ability to correctly
perform structural table
tasks under varying table
representation formats.
We design noise operators that emulate real
world table challenges (e.g., uninformative
sequential headers or merged cells) or
even adversarial behavior (e.g. shuffled or arbitrary column names).

\begin{figure*}[hbt]
    \centering
    \includegraphics[width=\textwidth]{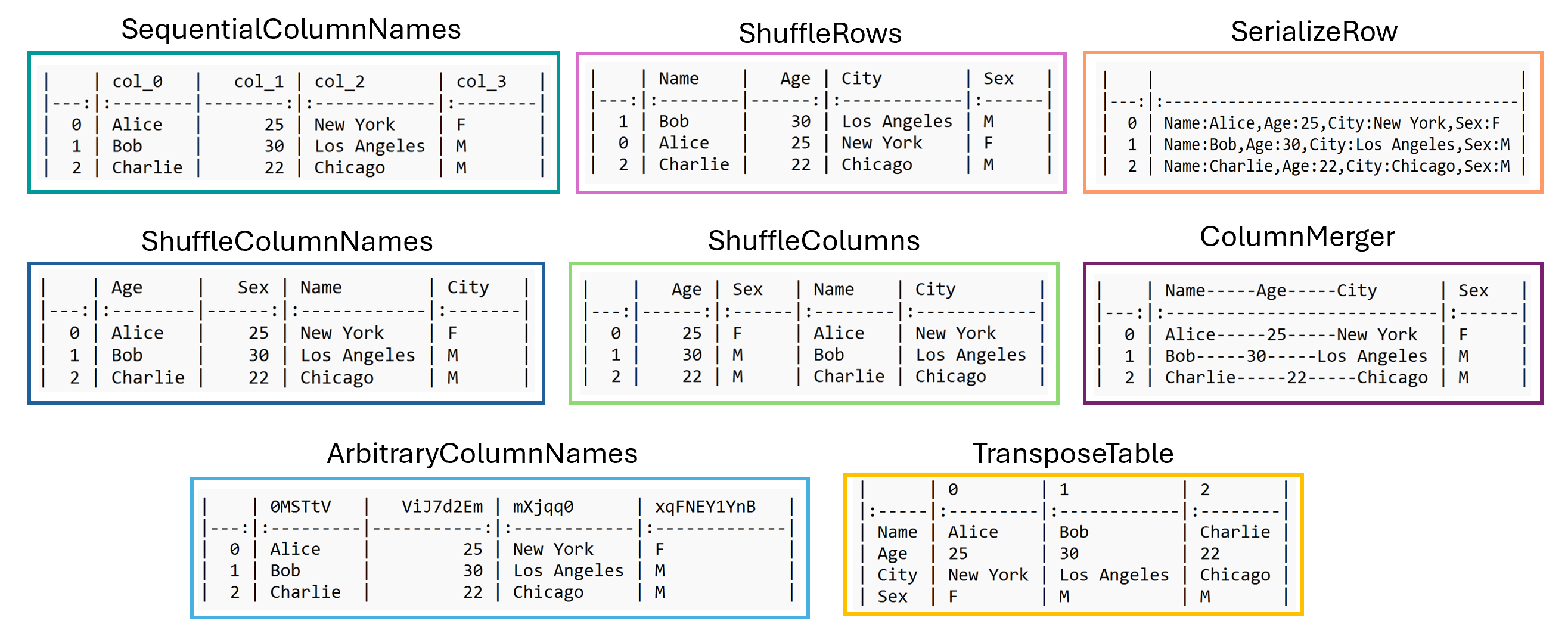}
    \caption{
    We apply eight different noise
    operations to
    test for the influence of spatial invariance,
    header rows information, 
    and
    the presence
    of semi-structured content
    on structural table
    task performance.
    }
    \label{fig:tabNoises}
\end{figure*}

\noindent{\textbf{Spatial Invariance}}: Tables often need to be rearranged or 
transformed
to be used. For example, 
long tables with many columns may need to be
transposed to faciliate plotting or for better readability.
Inspired by these challenges, we introduced
the following noise operations:
\begin{itemize}
    \item \textit{ShuffleRows}: We randomly reorder table rows.
    \item \textit{ShuffleColumns}: We randomize the order
    of columns within the table.
    \item \textit{TransposeTable}: We transpose the table.
\end{itemize}

\noindent{\textbf{Headers}}: Table headers
often play an important role in table understanding,
providing pseudo-natural language information about
their content and facilitating referencing.
However, in practice, user tables may not always
have informative or consistent
headers, or adversarial
actors may remove header
information altogether.
To simulate such cases
we introduce the following noise operations:
\begin{itemize}

    \item \textit{ArbitraryColumnNames}: We arbitrarily rename
    headers to randomly drawn alphanumeric sequences.

     \item \textit{SequentialColumnNames}: We rename headers
     to sequential entries of the form \texttt{col\_0, col\_1}, so on.

    \item \textit{ShuffleColumnNames}: We shuffle header names, while keeping data intact.
\end{itemize}

\noindent{\textbf{Semi-structured Content}}: 
Tables may contain columns that
have semi-structured content (e.g. phone numbers)
or users may need to start by parsing the table from
a semi-structured representation.
To induce such semi-structured data we use two noise operations:

\begin{itemize}

    \item \textit{SerializeRow}: We transform each row
    into a string of 
    key-value pairs. The resulting table has only one column in it. 
    \item \textit{ColumnMerger}: We merge randomly chosen 2, 3 and 4 contiguous columns together by
    adding (within each row) a \lstinline|-----| between their values.
    
\end{itemize}

Figure ~\ref{fig:tabNoises} shows each of these
noise operations applied to a table.

\subsection{Self-Supervised Structural Tasks}
We employ
self-supervised structural
tasks~\cite{sui2023evaluating},
which can be automatically generated,
to evaluate the extent to which
formats and noise operations
affect the LLM's ability to understand table structure.

\begin{figure*}[hbt]
    \centering
    \includegraphics[width=\textwidth]{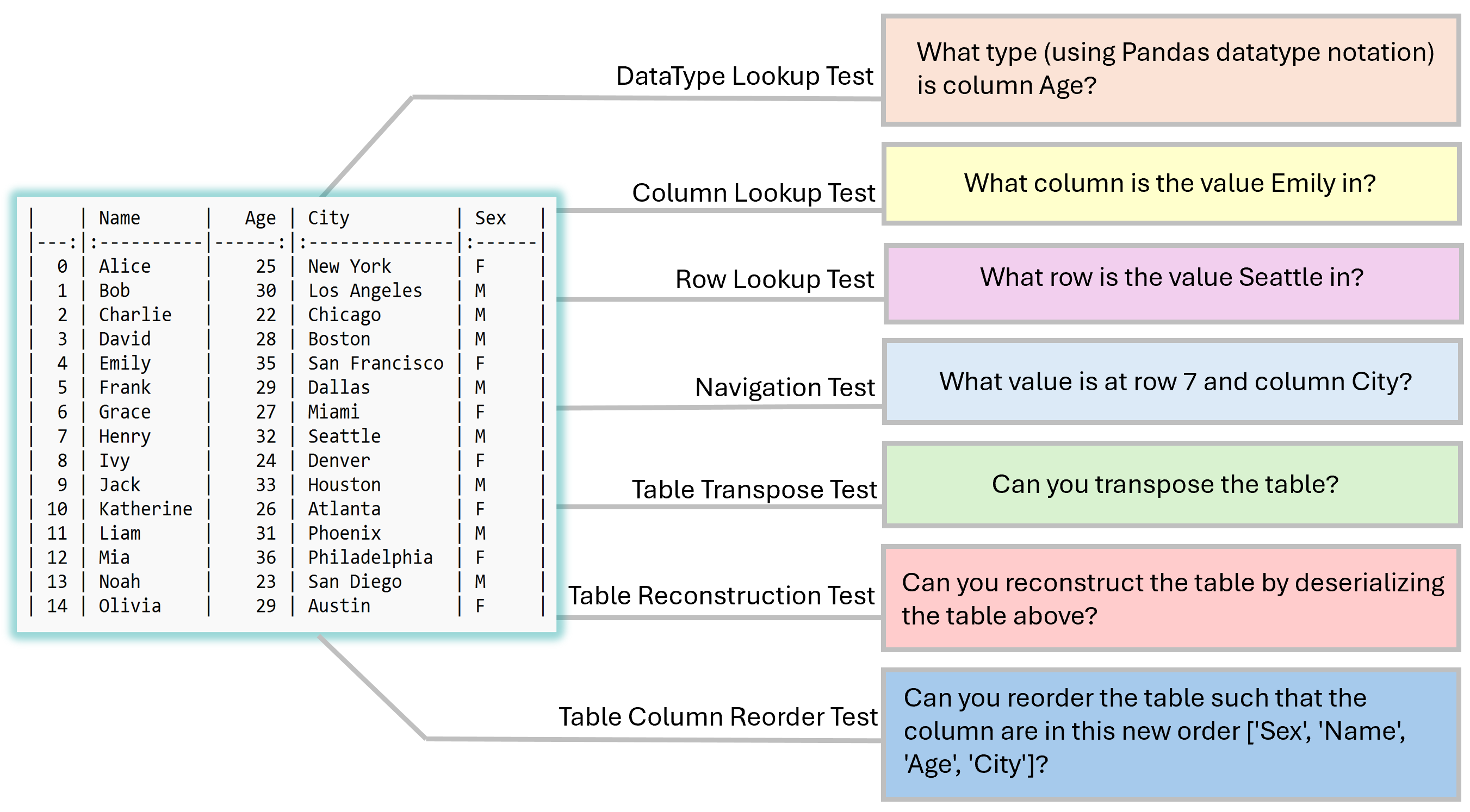}
    \caption{
    We generate self-supervised
    structural table understanding
    tasks: fact-finding tasks (e.g.
    navigation) and transformation tasks
    (e.g. table transposition).
}
    \label{fig:tab_TestCase_Examples}
\end{figure*}

We consider the following
structural fact-finding tasks:

 \begin{itemize}
     \item Navigation Test: Given 
    row and column coordinates, retrieve the value at that location.
     The model succeeds if it retrieves the value
     at those coordinates.

     \item Column Lookup Test: Given a value, retrieve 
     the name of a column that contains that value. 
     The model succeeds if it retrieves the
     name of a column that contains that value.
     
    \item Row Lookup Test: Given a value, retrieve the row index for a row that contains it. 
    The model succeeds if it retrieves the index
    of a row that contains that value.

    \item Data Type Lookup Test: Given a column name, determine
    the associated Pandas API datatype for the column values. The model succeeds if the datatype
    matches the groundtruth.

\end{itemize}

In addition to fact-finding
tasks, we introduce 
transformation tasks
that require manipulating the whole table:

\begin{itemize}
  \item Table Reconstruction Test: 
  Given a table, we serialize it (applying the Serialize Rows
  operation previously described and then joining
  rows with new line character). The model must parse the table and generate its output
  in one of our 8 table formats.

    \item Table Transpose Test:  Given a table,
    the model is tasked with transposing the table.

    \item Table Column Reordering Test:  Given 
    a table and a new (random) column order, the model
    must reorder the columns to match the indicated
    order. 
\end{itemize}

To measure the success of transformation tasks we compute 
precision, recall, and F1 score over the table values based on coordinates. 
Figure~\ref{fig:tab_TestCase_Examples} shows an
example for each
task. 
\section{Experimental Setup}
Our evaluation is designed to answer the following research questions:
\begin{itemize}
    
    \item [\textbf{RQ1}:] How does the table format impact LLM performance for self-supervised structural table understanding tasks?
    
    \item[\textbf{RQ2}:] How do noise operations impact LLM performance
    across different table formats?
    
\end{itemize}

\subsection{Experimental Setup}
We use
OpenAI's
GPT3 (\textit{text-davinci-003} endpoint)~\cite{brown2020language}.
Exploring cross-LLM behavior is left to future
work. We generate responses with temperature 0 to encourage deterministic behavior. Our prompts
have a token limit
of 4097, as determined by
the underlying LLM.
For each (table, format, noise operation, structural task)
we generate 100 tests\footnote{For HTML format we generate 50 tests per (table, format, noise operation, structural task),
due to token limits and throttling}
for fact-finding tasks and
25 tests for transformation tasks.
For each fact-finding test
we generate 15 completions and
for each transformation test
we generate 5 completions.

We report average performance metrics over tests.
For fact-finding tasks we compute pass@1~\cite{chen2021evaluating}
for each test. For transformation tasks we compute cell-wise
precision, recall and F1 per completion and average them. Computing exact table match, as needed for
pass@1 would not account for
partial performance. 
When reporting statistical significance, we perform comparisons using
the t-test (SciPy T-Test implementation) and perform
Bonferroni correction for multiple comparisons~\cite{bonferroni}.

We evaluate on 7 public datasets
collected from the popular data-science website Kaggle~\cite{kaggle}.
We chose datasets that are popularly used for both
classification and regression tasks:
AirQuality, HousingData, Diabetes, Wine Testing, Iris and Titanic.
We remove all rows where null values are present to avoid
creating spurious tasks.

\section{Results}

\subsection{RQ1: Impact of Formats on LLMs performance on Different Task}

\noindent{\textbf{Fact-Finding Tasks}}:
 Table~\ref{RQ1_a} summarizes pass@1 rates for different
 formats across our fact-finding tasks.
\begin{table}[tbh]
    \centering
    \tiny
    \caption{
    Average Pass@1
    for fact-finding tasks.
    DFLoader provides overall high pass@1 performance.
    } 
    \label{RQ1_a}
    
   \begin{tabular}{lrrrrrl}
    \toprule
        Table Formats & ColumnLookupTests & DataTypeLookupTests & NavigationTests & RowLookupTests  &Overall\\ \midrule
        
        \textsc{CommaSeparated} & 64.43 & 95.00 & 65.57 & 78.14  &75.78\\
        \textsc{DFloader} & 72.71 & 95.29 & 68.29 & 82.86  &\textbf{79.79}\\
        \textsc{DataMatrix} & 62.57 & 84.00 & 56.57 & \textbf{87.43}  &72.64\\
        \textsc{Json} & 65.00 & \textbf{96.43} & \textbf{71.43} & 78.86  &77.93\\
         \textsc{Markdown} & 61.43 & 85.86 & 48.71 & 73.29  &67.32\\
         \textsc{TabSeparated} & 67.00 & 94.00 & 64.43 & 78.14  &75.8\\
         \textsc{HTML} & \textbf{79.83} & 94.67 & 58.83 & 52.33  &71.4\\
         \textsc{HTMLNoSpace} & 73.00 & 93.50 & 62.00 & 59.50  &72.00\\
 \bottomrule
    \end{tabular}
\end{table}
We find that performance can vary substantially by format and
task. For example, while Markdown is a popular format for
data scientists sharing results, using this format for
tabular representation results in the
worst ColumnLookupTest performance --- 18.4\% points
lower than the best performing HTML format 
(p-value < $\frac{0.01}{7}$).

In contrast to prior work~\cite{sui2023evaluating}, we found that the HTML format underperforms alternatives like the JSON and
DFLoader formats. However, HTML did result in the highest
performance for one of our fact-finding tests: ColumnLookUpTest,
where the average pass@1 was 6.38\% higher than
the next best 
(p-value < $\frac{0.01}{7}$). 
A substantial
downside of HTML as a table representation is its verbosity:
in our experiments, using HTML results in
up to half as many rows being included compared to
other formats. Removing spaces in HTML improves
this slightly but the challenge remains.

JSON format, which is a popular serialization format, outperformed alternatives in the NavigationTests:
5.86\% higher than the Comma Separated format
(p-value < $\frac{0.01}{7}$).
We hypothesize that this
performance stems from a combination of orderly structure
and repeating navigation elements (specifically headers).
As shown in Figure~\ref{fig:tabformats}, every row is laid out in a separate line with
an associated key (showing the row index) and each row contains
a dictionary where keys are header names.

Our results further emphasize the brittleness of
LLMs to minor changes in structure representation. For example, while
there is relatively little difference between
the DataMatrix format and Comma Separated format,
RowLookup performance for Data Matrix was 9.3\%
higher 
(p-value < $\frac{0.01}{7}$).

On average, across our fact-finding tasks, we found that 
DFLoader format, which is essentially a code snippet in Pandas,
demonstrates competitive results and may be a suitable choice
for prompts where the user does not yet known what
kind of fact-finding knowledge is important for their task.

Finally, we find that all of our formats perform relatively
well in our DataTypeLookUpTests, highlighting that 
different table formats may not play a substantially role in
understanding the type of values (e.g. string versus numeric).

\noindent{\textbf{Transformation Tasks}}:
Our transformation tasks require that a format
be suitable for whole-table transformations.
Our results are summarized in Table~\ref{RQ1_c}.

Overall, we found that DFLoader and JSON format outperformed
alternatives for all the table transformation tasks. 
We hypothesize that this stems from isolation and repetition of key structural elements,
which enable 
use of local context to carry out whole-table tasks: DFLoader
presents each column in a separate list, and
JSON repeats headers locally.
For example, 
TableTranspose over our JSON format can effectively be carried
out per-line, compared to transposition over
a format like comma separated values, which requires
more complex retrievals (e.g. all header values are
in first row).

Similarly to the fact-finding tasks, 
Markdown format results in low performance across all our
tasks providing further evidence that such a format
should not be used for prompts for tabular data.
For example, Markdown's F1 score for column reordering is 49.67\% 
lower than JSON's (p-value < $\frac{0.01}{8}$).

\begin{table}[b]
    \centering
     \tiny
    \caption{
    F1 scores for transformation tasks. 
    DFLoader and JSON format,
    with structural element isolation and repetition,
    enable high performance on average
    across transformation
    tasks.}
    \label{RQ1_c}
    \begin{tabular}{lrrrr}
\toprule
      Table & TableColumnReorderTests      & TableReconstructionTests &      TableTransposeTests &  Overall\\   

\midrule
\textsc{CommaSeparated} & 95.33 & 74.33 & 99.00  &89.55
\\
\textsc{DFloader} & 99.33 & \textbf{98.00} & 98.33  &\textbf{98.55}
\\
\textsc{DataMatrix} & 92.67 & 90.67 & 0.00  &61.11
\\
\textsc{Json} & \textbf{99.67} & 85.00 & \textbf{100.00}  &94.89
\\
\textsc{Markdown} & 50.00 & 24.33 & 34.00  &36.11
\\
\textsc{TabSeparated} & 93.33 & 92.33 & 50.00  &78.55
\\
\textsc{HTML} & 50.00 & 86.00 & 83.33  &73.11
\\
\textsc{HTMLNoSpace} & 83.33 & 84.00 & 83.33  &83.55
\\
\bottomrule
\end{tabular}
\end{table}

\subsection{RQ2: Impact of Noise Operations on LLM's Performance on Structural Tasks. }

\noindent{\textbf{Fact-Finding Tasks}}: Table~\ref{RQ2_a_micro} presents
results of the impact
of different noise operations
on a subset of formats for our
fact-finding tasks,
chosen based on RQ1 performance.
The first takeaway from these 
experiments is that different
noise operations have a different
impact on formats and particular
fact-finding tasks. Furthermore,
this impact can be both positive
and negative.

For example, we find that 
transposing the input table and
representing it as JSON results in an
improvement of 
20.86\% 
(p-value < $\frac{0.01}{8}$)
at the navigation tests
compared to the original
input. However, this same transformation
substantially degrades
column and row lookup tests. After
inspecting generations, we found that
the LLM's generations for these tasks
seem to ignore the transposition and
often reply with the former headers (now row indices) as column names and viceversa.

For both the DataMatrix format
and HTML formats, we found that 
introducing noise into the
header names through operations like
shuffling column names, sequential column renaming, and arbitrary column renaming
resulted in degraded performance across
our navigation and column lookup tests. For example,
the Data Matrix format with sequential column naming resulted in
38\% 
(p-value < $\frac{0.01}{8}$) 
and
27.14\% 
(p-value < $\frac{0.01}{8}$) 
declines in 
navigation tests and column lookup tests, respectively.

Similarly, inducing semi-structured content
results can lower performance.
Serializing rows
results in worse performance
for data type detection across
our formats. For example, JSON
format's pass@1 score drops
by 12.43\%  
(p-value < $\frac{0.01}{8}$).
Merging cells impacts column lookup tests negatively, while
not impacting (or in some cases even improving) row lookup 
performance. For example, Data Matrix
format's pass@1 score drops
by 8\% 
(p-value < $\frac{0.01}{8}$)
in column lookup tests after applying the column merger noise operation.

\definecolor{lightestblue}{rgb}{0.9, 0.9, 1}
\definecolor{lightblue}{rgb}{0.7, 0.7, 1}
\definecolor{blue}{rgb}{0.4, 0.4, 1}
\definecolor{darkblue}{rgb}{0, 0, 0.7}

\definecolor{lightestgreen}{rgb}{0.9, 1, 0.9}
\definecolor{lightgreen}{rgb}{0.7, 1, 0.7}
\definecolor{green}{rgb}{0.4, 1, 0.4}
\definecolor{darkgreen}{rgb}{0.2, 0.8, 0.2}

\definecolor{lightestred}{rgb}{1, 0.8, 0.8}
\definecolor{lightred}{rgb}{1, 0.6, 0.6}
\definecolor{darkred}{rgb}{1, 0.4, 0.4}
\definecolor{darkestred}{rgb}{1, 0.2, 0.2}

\newcommand{\colorize}[1]{%
\ifdim #1 pt < 0 pt \relax
  \def\minvalue{0}
  \def\maxvalue{100}
\else
  \def\minvalue{0}
  \def\maxvalue{25}
\fi
  
  \pgfmathsetmacro{\firstquartile}{\minvalue + 0.25*(\maxvalue-\minvalue)}
  \pgfmathsetmacro{\secondquartile}{\minvalue + 0.50*(\maxvalue-\minvalue)}
  \pgfmathsetmacro{\thirdquartile}{\minvalue + 0.75*(\maxvalue-\minvalue)}

  \ifdim #1 pt < 0 pt \relax
    \ifdim #1 pt > -\firstquartile pt \relax
      \cellcolor{lightestred}
    \else
      \ifdim #1 pt > -\secondquartile pt \relax
        \cellcolor{lightred}
      \else
        \ifdim #1 pt > -\thirdquartile pt \relax
          \cellcolor{darkred}
        \else
          \cellcolor{darkestred}
        \fi
      \fi
    \fi
  \else
    \ifdim #1 pt < \firstquartile pt \relax
      \cellcolor{lightestgreen}
    \else
      \ifdim #1 pt < \secondquartile pt \relax
        \cellcolor{lightgreen}
      \else
        \ifdim #1 pt < \thirdquartile pt \relax
          \cellcolor{green}
        \else
          \cellcolor{darkgreen}
        \fi
      \fi
    \fi
  \fi
  #1}

\begin{table}[!htb]
\centering
    \tiny
    \caption{
    Average pass@1 delta from original
    to noisy for fact-finding tasks. Statistically significant values
    (p-value < $\frac{0.01}{8}$)
    are marked with "**". }
    \label{RQ2_a_micro}
\begin{tabular}{llrrrr}
\toprule
Table Format                           & Table Manipulation     & NavigationTests & ColumnLookupTests & RowLookupTests & DataTypeLookupTests \\ \midrule

\multirow{9}{*}{\textsc{Json}} & OriginalData & 71.43 & 65.00 & 78.86 & 96.43 \\
\cmidrule(l){2-6} 
 & ShuffleRows & \colorize{+0.57} & \colorize{+1.43} & \colorize{-6.57} & \colorize{+0.14} \\
 & ShuffleColumns & 0.00 & \colorize{+1.14} & \colorize{-6.72} & \colorize{-1.86} \\
 & ShuffleColumnNames & \colorize{-1.57} & \colorize{+1.43} & \colorize{-6.57} & \colorize{-8.86}** \\
 & SequentialColumnNames & \colorize{-1.72} & \colorize{+24.57}** & \colorize{-4.29} & \colorize{-1.01} \\
 & ArbitraryColumnNames & \colorize{-4.43} & \colorize{+23.14}** & \colorize{-10.43}** & \colorize{+0.57} \\
 & TransposeTable & \colorize{+20.86}** & \colorize{-65.00}** & \colorize{-76.29}** & \colorize{-33.86}** \\
 & ColumnMerger & \colorize{+7.28}** & \colorize{-7.71}** & \colorize{+2.28} & \colorize{-5.57}** \\
 & SerializeRow & \colorize{+16.57}** & \colorize{+3.14}** & \colorize{+2.57} & \colorize{-12.43}** \\
\midrule
\multirow{9}{*}{\textsc{DFloader}} & OriginalData & 68.29 & 72.71 & 82.86 & 95.29 \\
\cmidrule(l){2-6} 
 & ShuffleRows & \colorize{-23.29}** & \colorize{-4.85} & \colorize{-44.57}** & \colorize{+1.42} \\
 & ShuffleColumns & \colorize{+2.42} & \colorize{+4.72} & \colorize{-0.86} & \colorize{+2.85} \\
 & ShuffleColumnNames & \colorize{+3.14} & \colorize{-3.57} & \colorize{-10.43}** & \colorize{-3.86}** \\
 & SequentialColumnNames & \colorize{+2.28} & \colorize{+11.43}** & \colorize{-10.72}** & 0.00 \\
 & ArbitraryColumnNames & \colorize{+2.14} & \colorize{+0.58} & \colorize{-11.29}** & \colorize{+1.28} \\
 & TransposeTable & \colorize{+3.00} & \colorize{-52.85}** & \colorize{-69.29}** & \colorize{-29.43}** \\
 & ColumnMerger & \colorize{-7.72}** & \colorize{-4.57} & \colorize{-3.43} & \colorize{-2.15} \\
 & SerializeRow & \colorize{-3.86} & \colorize{+10.58}** & \colorize{-15.86}** & \colorize{-16.00}** \\
\midrule
\multirow{9}{*}{\textsc{DataMatrix}} & OriginalData & 56.57 & 62.57 & 87.43 & 84.00 \\
\cmidrule(l){2-6} 
 & ShuffleRows & \colorize{-17.43}** & \colorize{-4.00} & \colorize{-31.72}** & \colorize{+2.29} \\
 & ShuffleColumns & \colorize{-6.57} & \colorize{-2.14} & \colorize{-0.72} & \colorize{+1.12} \\
 & ShuffleColumnNames & \colorize{-20.57}** & \colorize{-23.14}** & \colorize{-1.86} & \colorize{-15.00}** \\
 & SequentialColumnNames & \colorize{-38.00}** & \colorize{-27.14}** & \colorize{+2.28} & \colorize{-7.43}** \\
 & ArbitraryColumnNames & \colorize{-17.71}** & \colorize{-23.71}** & \colorize{-1.57} & \colorize{-2.43} \\
 & TransposeTable & \colorize{-4.57} & \colorize{-60.00}** & \colorize{-85.14}** & \colorize{-22.00}** \\
 & ColumnMerger & \colorize{-10.71}** & \colorize{-8.00}** & \colorize{-2.00} & \colorize{+4.86} \\
 & SerializeRow & \colorize{-22.57}** & \colorize{+8.72}** & \colorize{-39.57}** & \colorize{-1.00} \\
\midrule
\multirow{9}{*}{\textsc{HTML}} & OriginalData & 58.83 & 79.83 & 52.33 & 94.67 \\
\cmidrule(l){2-6} 
 & ShuffleRows & \colorize{-1.50} & \colorize{-1.50} & \colorize{-3.50} & \colorize{+1.83} \\
 & ShuffleColumns & \colorize{-1.16} & \colorize{-2.33} & \colorize{+4.00} & \colorize{-0.34} \\
 & ShuffleColumnNames & \colorize{-20.50}** & \colorize{-27.16}** & \colorize{-11.66}** & \colorize{-15.67}** \\
 & SequentialColumnNames & \colorize{-27.66}** & \colorize{-36.16}** & \colorize{+9.34}** & \colorize{-19.67}** \\
 & ArbitraryColumnNames & \colorize{-12.16}** & \colorize{-30.58} & \colorize{-7.33} & \colorize{-2.00} \\
 & TransposeTable & \colorize{-25.08} & \colorize{-79.83} & \colorize{-9.83} & \colorize{-49.92} \\
 & ColumnMerger & \colorize{+13.17}** & \colorize{-25.83}** & \colorize{+3.17} & \colorize{-6.34}** \\
 & SerializeRow & \colorize{+24.84}** & \colorize{+3.50} & \colorize{-1.33} & \colorize{-18.67}** \\

 \bottomrule 
\end{tabular}
\end{table}

\noindent{\textbf{Transformation Tasks}}: Table~\ref{tab:RQ2_delta_macro}
presents our transformation task results
after applying noise operations.
We discuss multiple interesting trends.

First, we find that introducing
sequence information into headers
(through the sequential column renaming
noise operation) can significantly impact
performance for the column reordering
task (which requires changing column order)
for some formats. For example,
for the comma separated format, introducing
sequential column renaming degrades
column reordering F1-score by
67.33\%
(p-value < $\frac{0.01}{8}$).
Column name shuffling and arbitrary
column renaming, which \emph{do not} introduce
any form of sequential bias reduce performance
as well, but by a smaller margin.

Second, we find that table transpose performance
can be significantly affected by \emph{transposing} the table initially.
For example, transposing the
table in JSON reduces the
transpose task F1-score by 
89\%
(p-value < $\frac{0.01}{8}$).
This emphasizes that preprocessing may
be necessary for tabular data, compared to
relying on the model to perform
such transformations itself for downstream
tasks. 

Finally, we find that introducing unstructured
content can impact transformation tasks.
For example, we find that JSON format, which obtains high table transposition performance, drops to zero (p-value < $\frac{0.01}{8}$), when
the column merging noise operation is applied.

\begin{table}[]
    \centering
    \tiny
    \caption{
    Average F1 score delta from original
    to noisy for transformation tasks. Statistically significant values
    (p-value < $\frac{0.01}{8}$)
    are marked with "**".}

    \label{tab:RQ2_delta_macro}
    \begin{tabular}{llrrr}
\toprule
     Table Formats& Table Manipulation & TableColumnReorderTests               & TableReconstructionTests &    TableTransposeTests            \\
\midrule
\multirow{9}{*}{\textsc{Json}} & OriginalData & 99.67 & 85.00 & 100.00 \\
\cmidrule(l){2-5} 
 & ShuffleRows & \colorize{-1.00} & \colorize{-45.00}\mbox{$^{**}$} & \colorize{-13.33}** \\
 & ShuffleColumns & \colorize{+0.33} & \colorize{-19.00} & \colorize{-40.67}** \\
 & ShuffleColumnNames & \colorize{-0.34} & \colorize{-13.67} & \colorize{-29.33}** \\
 & SequentialColumnNames & \colorize{+0.33} & \colorize{-9.00} & \colorize{-2.00} \\
 & ArbitraryColumnNames & \colorize{+0.33} & \colorize{-4.33} & \colorize{-0.67} \\
 & TransposeTable & \colorize{-89.00}** & \colorize{-78.33}** & \colorize{-42.00}** \\
 & ColumnMerger & \colorize{+0.33} & \colorize{-85.00}** & \colorize{-75.33}** \\
 & SerializeRow & \colorize{-59.00}** & \colorize{-46.33}** & \colorize{-100.00}** \\
\midrule
\multirow{9}{*}{\textsc{DFloader}} & OriginalData & 99.33 & 98.00 & 98.33 \\
\cmidrule(l){2-5} 
 & ShuffleRows & \colorize{+0.67} & \colorize{-78.67}** & \colorize{-16.33}** \\
 & ShuffleColumns & \colorize{+0.67} & \colorize{-34.00}** & \colorize{-34.33}** \\
 & ShuffleColumnNames & \colorize{+0.67} & \colorize{-31.33}** & \colorize{-26.33}** \\
 & SequentialColumnNames & \colorize{+0.67} & \colorize{-54.67}** & \colorize{-1.00} \\
 & ArbitraryColumnNames & \colorize{+0.67} & \colorize{-18.00}** & \colorize{-0.33} \\
 & TransposeTable & \colorize{-43.33}** & \colorize{-98.00}** & \colorize{-83.00}** \\
 & ColumnMerger & \colorize{+0.67} & \colorize{-98.00}** & \colorize{-81.00}** \\
 & SerializeRow & \colorize{-33.33}** & \colorize{-73.33}** & \colorize{-60.33}** \\
\midrule
\multirow{9}{*}{\textsc{CommaSeparated}} & OriginalData & 95.33 & 74.33 & 99.00 \\
\cmidrule(l){2-5} 
 & ShuffleRows & \colorize{-7.33} & \colorize{-41.66}** & \colorize{-70.33}** \\
 & ShuffleColumns & \colorize{-4.66} & \colorize{-19.00} & \colorize{-33.00}** \\
 & ShuffleColumnNames & \colorize{-32.00}** & \colorize{-9.66} & \colorize{-47.00}** \\
 & SequentialColumnNames & \colorize{-67.33}** & \colorize{-13.00} & \colorize{-24.33}** \\
 & ArbitraryColumnNames & \colorize{-28.66}** & \colorize{+4.34} & \colorize{-21.67}** \\
 & TransposeTable & \colorize{+2.00} & \colorize{-65.00}** & \colorize{-98.33}** \\
 & ColumnMerger & \colorize{-4.66} & \colorize{-74.33}** & \colorize{-80.33}** \\
 & SerializeRow & \colorize{-95.33}** & \colorize{-57.66}** & \colorize{-99.00}** \\
 \midrule
 \multirow{9}{*}{\textsc{TabSeparated}} & OriginalData & 93.33 & 92.33 & 50.00 \\
 \cmidrule(l){2-5} 
 & ShuffleRows & \colorize{-4.00} & \colorize{-57.00}** & \colorize{-34.67}** \\
 & ShuffleColumns & \colorize{-6.00} & \colorize{-31.00}** & \colorize{-6.00}** \\
 & ShuffleColumnNames & \colorize{-59.33}** & \colorize{-29.66}** & 0.00 \\
 & SequentialColumnNames & \colorize{-68.00}** & \colorize{-27.00}** & \colorize{-7.33}** \\
 & ArbitraryColumnNames & \colorize{-45.33}** & \colorize{-13.00}** & \colorize{-2.00}** \\
 & TransposeTable & \colorize{-44.66}** & \colorize{-83.00}** & \colorize{-50.00}** \\
 & ColumnMerger & \colorize{-41.33}** & \colorize{-92.33}** & \colorize{-48.00}** \\
 & SerializeRow & \colorize{-93.33}** & \colorize{-91.66}** & \colorize{-50.00}** \\

\bottomrule
\end{tabular}
\end{table}

\section{Conclusion}
We evaluated LLM performance
on self-supervised 
structural table understanding tasks using
different formats and noise operations.
Our results show that different formats
obtain varying performance and noise
operations can change results (both positively
and negatively). Future work should
consider cross-LLM performance,
further exploring what format
properties correlate with performance,
and evaluating whether performance on 
table structure understanding tasks
correlates with performance on downstream 
table task such as question answering
or NL-to-code generation.


\clearpage
\begin{thebibliography}{10}

\bibitem{bonferroni}
C.~Bonferroni.
\newblock Teoria statistica delle classi e calcolo delle probabilita.
\newblock {\em Pubblicazioni del R Istituto Superiore di Scienze Economiche e Commericiali di Firenze}, 8:3--62, 1936.

\bibitem{brown2020language}
T.~Brown, B.~Mann, N.~Ryder, M.~Subbiah, J.~D. Kaplan, P.~Dhariwal, A.~Neelakantan, P.~Shyam, G.~Sastry, A.~Askell, et~al.
\newblock Language models are few-shot learners.
\newblock {\em Advances in neural information processing systems}, 33:1877--1901, 2020.

\bibitem{cahoon2022need}
J.~Cahoon, A.~Savelieva, A.~C. Mueller, A.~Floratou, C.~Curino, H.~Patel, J.~Henkel, M.~Weimer, N.~Gustafsson, R.~Wydrowski, et~al.
\newblock The need for tabular representation learning: An industry perspective.
\newblock In {\em NeurIPS 2022 First Table Representation Workshop}, 2022.

\bibitem{chen2021evaluating}
M.~Chen, J.~Tworek, H.~Jun, Q.~Yuan, H.~P. d.~O. Pinto, J.~Kaplan, H.~Edwards, Y.~Burda, N.~Joseph, G.~Brockman, et~al.
\newblock Evaluating large language models trained on code.
\newblock {\em arXiv preprint arXiv:2107.03374}, 2021.

\bibitem{chen2022large}
W.~Chen.
\newblock Large language models are few (1)-shot table reasoners.
\newblock {\em arXiv preprint arXiv:2210.06710}, 2022.

\bibitem{chen2023large}
W.~Chen.
\newblock Large language models are few(1)-shot table reasoners, 2023.

\bibitem{herzig2020tapas}
J.~Herzig, P.~K. Nowak, T.~M{\"u}ller, F.~Piccinno, and J.~M. Eisenschlos.
\newblock Tapas: Weakly supervised table parsing via pre-training.
\newblock {\em arXiv preprint arXiv:2004.02349}, 2020.

\bibitem{semantictypes:2023}
G.~Jaimovitch-L{\'o}pez, C.~Ferri, J.~Hern{\'a}ndez-Orallo, F.~Mart{\'\i}nez-Plumed, and M.~J. Ram{\'\i}rez-Quintana.
\newblock Can language models automate data wrangling?
\newblock {\em Machine Learning}, 112(6):2053--2082, 2023.

\bibitem{kaggle}
{Kaggle}.
\newblock {Kaggle}.
\newblock \url{https://www.kaggle.com}, 2023.
\newblock Accessed on 2023-10-02.

\bibitem{koleva2022analysis}
A.~Koleva, M.~Ringsquandl, and V.~Tresp.
\newblock Analysis of the attention in tabular language models.
\newblock In {\em NeurIPS 2022 First Table Representation Workshop}, 2022.

\bibitem{microsoft-data-cleaning}
{Microsoft}.
\newblock Top ten ways to clean your data, 2023.
\newblock Accessed on 2023-10-02.

\bibitem{Suhara:2022}
Y.~Suhara, J.~Li, Y.~Li, D.~Zhang, c.~Demiralp, C.~Chen, and W.-C. Tan.
\newblock Annotating columns with pre-trained language models.
\newblock In {\em Proceedings of the 2022 International Conference on Management of Data}, SIGMOD '22, page 1493–1503, New York, NY, USA, 2022. Association for Computing Machinery.

\bibitem{sui2023evaluating}
Y.~Sui, M.~Zhou, M.~Zhou, S.~Han, and D.~Zhang.
\newblock Evaluating and enhancing structural understanding capabilities of large language models on tables via input designs, 2023.

\bibitem{wang2021tuta}
Z.~Wang, H.~Dong, R.~Jia, J.~Li, Z.~Fu, S.~Han, and D.~Zhang.
\newblock Tuta: Tree-based transformers for generally structured table pre-training.
\newblock In {\em Proceedings of the 27th ACM SIGKDD Conference on Knowledge Discovery \& Data Mining}, pages 1780--1790, 2021.

\bibitem{wei2022chain}
J.~Wei, X.~Wang, D.~Schuurmans, M.~Bosma, F.~Xia, E.~Chi, Q.~V. Le, D.~Zhou, et~al.
\newblock Chain-of-thought prompting elicits reasoning in large language models.
\newblock {\em Advances in Neural Information Processing Systems}, 35:24824--24837, 2022.

\bibitem{zha2023tablegpt}
L.~Zha, J.~Zhou, L.~Li, R.~Wang, Q.~Huang, S.~Yang, J.~Yuan, C.~Su, X.~Li, A.~Su, et~al.
\newblock Tablegpt: Towards unifying tables, nature language and commands into one gpt.
\newblock {\em arXiv preprint arXiv:2307.08674}, 2023.

\end{thebibliography}
\end{document}